\documentclass[sensors,article,accept,moreauthors,pdftex]{Definitions/mdpi}

\firstpage{1}
\pubvolume{19}
\issuenum{24}
\articlenumber{5361}
\pubyear{2019}
\copyrightyear{2019}
\history{Received: 25 October 2019; Accepted: 29 November 2019; Published: 5 December 2019}
\updates{yes} 

\Title{Waterfall Atrous Spatial Pooling Architecture for Efficient Semantic Segmentation}

\Author{Bruno Artacho and Andreas Savakis * }

\AuthorNames{Bruno Artacho and Andreas Savakis}

\address[1]{%
Department of Computer Engineering, Rochester Institute of Technology, Rochester, NY 14623, USA;
bmartacho@mail.rit.edu}  

\corres{\hangafter=1 \hangindent=1.05em \hspace{-0.82em} Correspondence: andreas.savakis@rit.edu}

\abstract{
We propose a new efficient architecture for semantic segmentation, based on a ``Waterfall'' Atrous Spatial Pooling architecture, that achieves a considerable accuracy increase while decreasing the number of network parameters and memory footprint. The proposed Waterfall architecture leverages the efficiency of progressive filtering in the cascade architecture while maintaining multiscale fields-of-view comparable to spatial pyramid configurations.
Additionally, our method does not rely on a postprocessing stage with Conditional Random Fields, which further reduces complexity and required training time.
We demonstrate that the Waterfall approach with a ResNet backbone is a robust and efficient architecture for semantic segmentation obtaining state-of-the-art results with significant reduction in the number of parameters for the Pascal VOC dataset and the Cityscapes dataset.}


\begin{document}

\section{Introduction}
Semantic segmentation is an important computer vision task~\cite{Review, Review2, Review3} with applications in autonomous driving~\cite{Driving}, human--machine interaction~\cite{HMI}, computational photography~\cite{photo}, and image search engines~\cite{ImageSearch}.
The significance of semantic segmentation, in both the development of novel architectures and its practical use, has motivated the development of several approaches that aim to improve the encouraging initial results of Fully Convolutional Networks (FCN)~\cite{FCN}.
One important challenge to address is the decrease of the feature map size due to pooling, which requires unpooling to perform pixel-wise labeling of the image for segmentation.

DeepLab~\cite{DeepLab}, for instance, used dilated or Atrous Convolutions to tackle the limitations posed by the loss of resolution inherited from unpooling operations. The advantage of Atrous Convolution is that it maintains the Field-of-View (FOV) at each layer of the network. DeepLab implemented Atrous Spatial Pyramid Pooling (ASPP) blocks in the segmentation network, allowing the utilization of several Atrous Convolutions at different dilation rates for a larger FOV.

A limitation of the ASPP architecture is that the network experiences a significant increase in size and memory required. This limitation was addressed in~\cite{Rethinking}, by replacing ASPP modules with the application of Atrous Convolutions in series, or cascade, with progressive rates of dilation. However, although this approach successfully decreased the size of the network, it presented the setback of decreasing the size of the FOV.

\begin{figure}[H]
\centering
\includegraphics[width=1\linewidth]{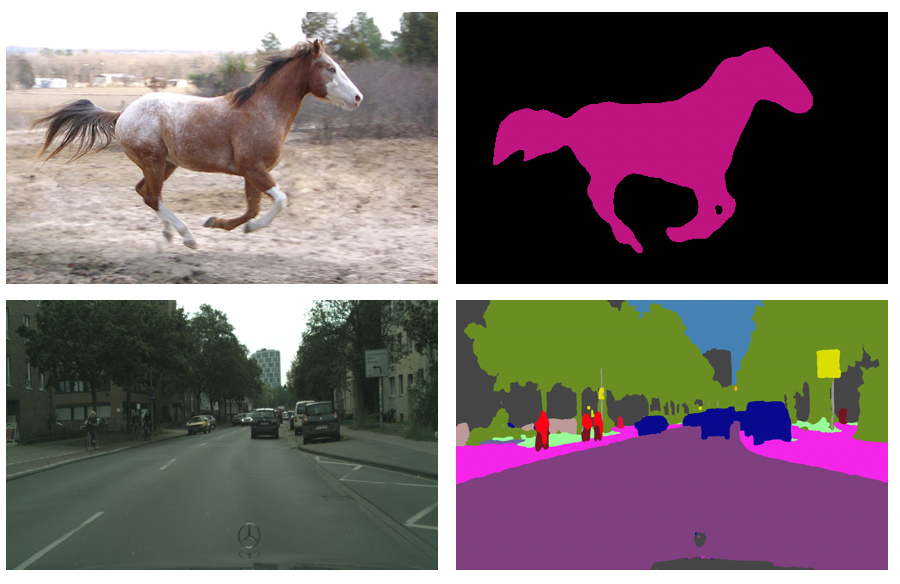}
\caption{Semantic segmentation examples using WASPnet.}
\label{fig:WASPnet_examples}
\end{figure}

Motivated by the success achieved by a network architecture with parallel branches introduced by the Res2Net module~\cite{Res2Net}, we incorporate Res2Net blocks in a semantic segmentation network. Then, we propose a novel architecture named the Waterfall Atrous Spatial Pooling (WASP) and use it in a semantic segmentation network we refer to as WASPnet (see segmentation examples in Figure \ref{fig:WASPnet_examples}).
Our WASP module combines the cascaded approach used in~\cite{Rethinking} for Atrous Convolutions with the larger FOV obtained from traditional ASPP in DeepLab for the deconvolutional stages of semantic~segmentation.

The WASP approach leverages the progressive extraction of larger FOV from cascade methods, and is able to achieve parallelism of branches with different FOV rates while maintaining reduced parameter size. The resulting architecture has a flow that resembles a waterfall, which is how it gets its~name.

The main contributions of this paper are as follows.
\begin{itemize}[leftmargin=*,labelsep=5.8mm]
\item We propose the Waterfall method for Atrous Spatial Pooling that achieves significant reduction in the number of parameters in our semantic segmentation network compared to current methods based on the spatial pyramid architecture.
\item Our approach increases the receptive field of the network by combining the benefits of cascade Atrous Convolutions with multiple fields-of-view in a parallel architecture inspired by the spatial pyramid approach.
\item Our results show that the Waterfall approach achieves  state-of-the-art  accuracy  with a significant reduction in the number of network parameters.
\item Due to the superior performance of the WASP architecture, our network does not require postprocessing of the semantic segmentation result with a CRF module, making it even more efficient in terms of computational complexity.
\end{itemize}

\section{Related Work}

The innovations in Convolutional Neural Networks (CNNs) by the authors of~\cite{AlexNet,VGG,GoogleNet,ResNet} form the core of image classification and serve as the structural backbone for state-of-the-art methods in semantic segmentation. However, an important challenge with incorporating CNN layers in segmentation is the significant reduction of resolution caused by pooling.

\begin{figure}[H]
\centering
\includegraphics[width=0.8\linewidth]{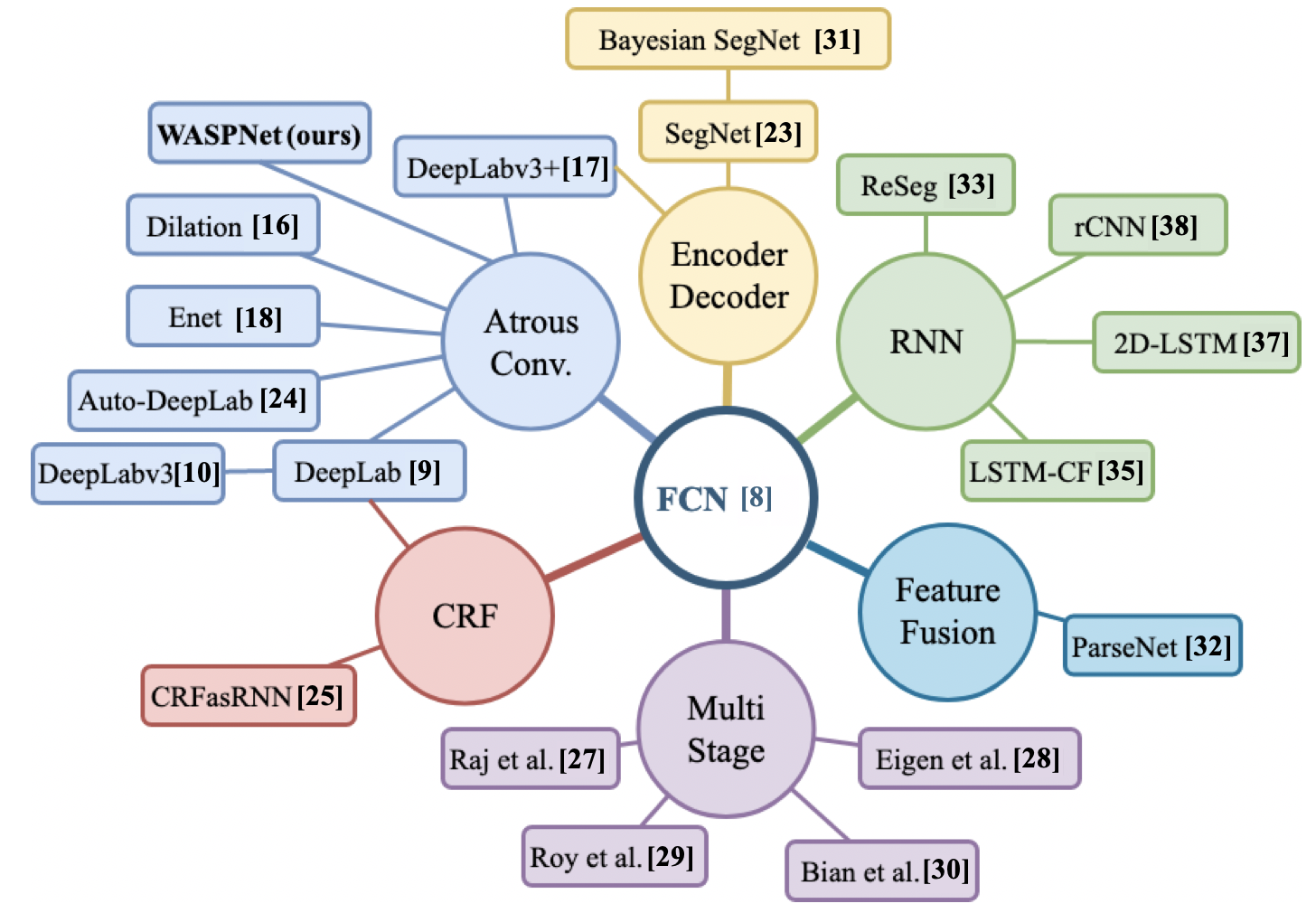}
\caption{Semantic segmentation research overview.}   
\label{fig:Segmentation}
\end{figure}

The breakthrough work of Long et al.~\cite{FCN} introduced Fully Convolutional Networks (FCN) by replacing the final fully connected layers with deconvolutional stages.
FCN~\cite{FCN} addressed the resolution reduction problem by deploying upsampling strategies across deconvolution layers.
These deconvolution stages attempt to reverse the convolution operation and increase the feature map size back to the dimensions of the original image.
The contributions of FCN~\cite{FCN} triggered research in semantic segmentation that led to a variety of different approaches that are visually illustrated in Figure \ref{fig:Segmentation}.

\subsection{Atrous Convolution}
The most popular technique shared among semantic segmentation architectures is the use of dilated or Atrous Convolutions. An early work by Yu et al.~\cite{DilatedConv} highlighted the uses of dilation.
Atrous convolutions were further explored by the authors of~\cite{DeepLab,Rethinking,DeepLabv3+,Enet}. The main objectives of Atrous Convolutions are to increase the size of the receptive fields in the network, avoid downsampling, and generate a multiscale framework for segmentation.

\begin{figure}[H]
\centering
\includegraphics[width=0.8\linewidth]{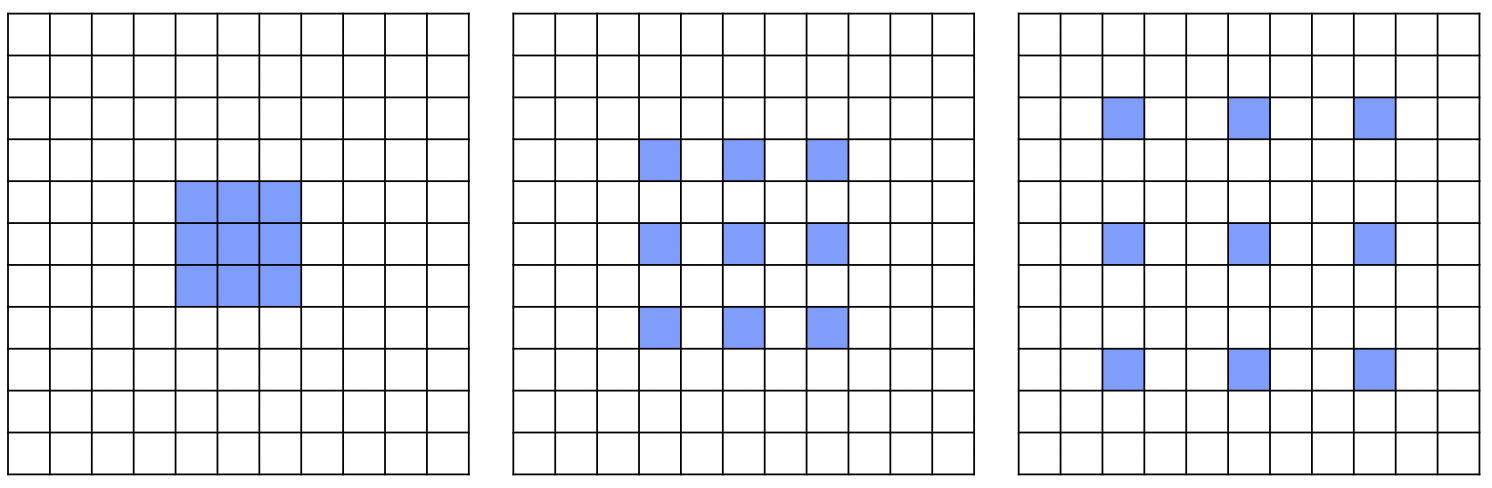}
\caption{Input pixels using a 3 $\times$ 3 Atrous Convolutios with different dilation rates of 1, 2, and 3,~respectively.}
\label{fig:Atrous}
\end{figure}

The name Atrous is derived from the French expression ``algorithm \`a trous'', or translated to English ``Algorithm with holes''. As alluded by its name, Atrous Convolutions alter the convolutional filters by the insertion of ``holes'', or zero values in the filter, resulting in the increased size of the receptive field, resembling a hybrid of convolution and pooling layers.
The use of Atrous Convolutions in the network is shown in Figure \ref{fig:Atrous}.

In the simpler case of a one-dimensional convolution, the output of the signal is defined as follows~\cite{DeepLab},
\begin{equation}
y[i]=\sum_{k=1}^{K}x[i+rk]\cdot w[k]
\end{equation}
where $r$ is the rate at which the Atrous Convolution is dilated, $\omega[k]$ is the filter of length K, $x[i]$ is the input, and $y[i]$ is the output of a pixel. As pointed out in~\cite{DeepLab}, a rate value of the unit results in a regular convolution operation.

Leveraging the success of the Spatial Pyramid Pooling (SPP) structure by He et al.~\cite{SPP}, the ASPP architecture was introduced in DeepLab~\cite{DeepLab}.
The special configuration of ASPP assembles dilated convolutions in four parallel branches with different rates. The resulting feature maps are combined by fast bilinear interpolation with an additional factor of eight to recover the feature maps in the original~resolution.

\subsection{DeepLabv3}
The application of Atrous Convolution followed the ASPP approach in~\cite{DeepLab} was later extended in~\cite{Rethinking} to the cascade approach, that is, the use of several Atrous Convolutions in sequence with rates increasing through its flux.
This approach, named Deeplabv3~\cite{Rethinking}, allows the architecture to perform deeper analysis and increment its performance using approaches similar to those in~\cite{Deformable}.

Contributions in~\cite{Rethinking} included module realization in a cascade fashion, investigation of different multi-grid configurations for dilation in the cascade of convolutions, training with different output stride scales for the Atrous Convolutions, and techniques to improve the results when testing and fine-tuning for segmentation challenges. Another addition presented by~\cite{Rethinking} is the inclusion of a ResNet101 model, pretrained on both ImageNet~\cite{ImageNet} and JFT-300M~\cite{Revisiting} datasets.

More recently, DeepLabv3+~\cite{DeepLabv3+} proposed the incorporation of ASPP modules with the encoder--decoder structure adopted by~\cite{Segnet}, reporting a better refinement in the border of the objects being segmented. This novel approach represented a significant improvement in accuracy from previous methods. In a separate development, Auto-DeepLab~\cite{Auto-DeepLab} uses an Auto-ML approach to learn a semantic segmentation architecture by searching both the network level and the cell level of the structure. It achieves results comparable to current methods without requiring ImageNet~\cite{ImageNet} pre-training or hierarchical architecture search.

\subsection{CRF}
A complication resulting of the lack of pooling layers is a reduction of spatial invariance.
Thus, additional techniques are used to recover  spatial definition, namely, Conditional Random Fields (CRF) and Atrous Convolutions. One popular method relying on CRF is CRFasRNN~\cite{CRFasRNN}. Aiming to better delineate objects in the image, CRFasRNN combines CNN and CRF in a single network to incorporate the probabilistic method of the Gaussian pairwise potentials during inference. That enables  end-to-end training, avoiding the need of postprocessing with a separate CRF module, as done in~\cite{DeepLab}. A limitation of architectures using CRF is that CRF has difficulty capturing delicate boundaries, as they have low confidence in the unary term of the CRF energy function.

The postprocessing module of CRF performs refining of the prediction by Gaussian filters and iterative comparisons of pixels in the output image. The iteration process aims to minimize the ``energy'' $E(x)$ below.
\begin{equation}
E(x)=\sum_{i}\theta_{i}(x_i)+\sum_{ij}\theta_{ij}(x_i,x_{j})
\end{equation}

The energy consists of the summations of the unary potentials $\theta_i(x_i) = -logP(x_i)$, where $P(x_i)$ is the probability (softmax) that pixel $i$ is correctly computed by the CNN, and the pairwise potential energy $\theta_{ij}(x_i,x_{j})$, which is determined by the relationship between two pixels.
Following the authors of~\cite{CRF}, $\theta_{ij}(x_i,x_j)$ is defined as
\begin{equation}
\theta_{ij}(x_i,x_{j})=\mu(x_i,x_j)\Big[\omega_1\cdot exp\Big(-\frac{||p_i-p_j||^2}{2\sigma_{\alpha}^{2}}-\frac{||I_i-I_j||^2}{2\sigma_{\beta}^{2}}\Big) +\omega_2\cdot exp\Big(-\frac{||p_i-p_j||^2}{2\sigma_{\gamma}^{2}}\Big)\Big]
\end{equation}
where the function $\mu(x_i,x_j)$ is defined to be equal to 1 in the case of $x_i \neq x_j$ and zero otherwise, that is, the CRF only accounts for energy that needs to be minimized when the labels differ. The pairwise potential function utilizes two Gaussian kernels: the first depends on pixel positions $p$ and the RGB color $I$; the second depends only on pixel positions. The Gaussian kernels are controlled by the hyperparameters $\sigma_{\alpha}$, $\sigma_{\beta}$, and  $\sigma_{\gamma}$, which are determined through the iterations of the CRF, as well as the weights $\omega_1$ and $\omega_2$.

\subsection{Other Methods}
In contrast to the large scale of segmentation networks using Atrous Convolutions, the Efficient Neural Network (ENet)~\cite{Enet} produces a real-time segmentation by trading-off some of its accuracy for a significant reduction in processing time, ENet is up to 18$\times$ faster than other architectures.

During learning, CNN architectures have the tendency to learn information that is specific to the scale of the input image dataset. In an attempt to deal with this issue, a multiscale approach is used. For instance, the authors of~\cite{Multi-scale-Raj} proposed a network with two paths containing the original resolution image and another with double the resolution. The former is processed through a short CNN and the latter through a fully convolutional VGG-16. The first path is then combined with the upsampled version of the second resulting in a network that can deal with larger variations in scale. \mbox{A similar} approach is applied in~\cite{Multi-scale-Eigen,Multi-scale-Roy,Multi-scale-Bian}, expanding the structure to include a larger amount of networks and~scales.

Other architectures achieved good results in semantic segmentation by using an encoder--decoder variant. For instance, SegNet~\cite{Segnet} utilizes both an encoder and decoder phase, while relying on pooling indices from the encoder phase to aid the decoder phase. The Softmax classifier generates the final segmentation prediction map. The architecture presented by SegNet was further developed to include Bayesian techniques to model uncertainty in the network~\cite{Bayesian-SegNet}.

Contrasting with the work in~\cite{FCN}, ParseNet~\cite{Parsenet} completes an early fusion in the network, by performing an early merge of the global features from previous layers with the current map of the posterior layer. In ParseNet, the previous layer is unpooled and concatenated to the following layers to generate the final classifier prediction with both having the same size. This approach differs from FCN where the skip connection concatenates maps of different sizes.

Recurrent Neural Networks (RNN) have been used to successfully combine pixel-level information with local region information, enabling the RNN to include global context in the construction of the segmented image. A limitation of RNN, when used for Semantic Segmentation, is~that it has difficulty constructing a sequence based on the structure of natural images. ReSeg~\cite{ReSeg} is a network based on previous work by ReNet~\cite{ReNet}. ReSeg presents an approach where RNN blocks from ReNet are applied after a few layers of a VGG structure, generating the final segmentation map by the use of upsampling by transposed convolutions. However, RNN-based architectures suffer from the vanishing gradient problem.

Networks using Long Short-Term Memory (LSTM) aim to tackle the issue of vanishing gradients. For instance, LSTM Context Fusion (LSTM-CF)~\cite{LSTM-CF} utilizes the concatenation of an architecture similar to DeepLab to process RGB and depth information. It uses three different scales for the RGB feature response and depth, similar to the work in~\cite{DeepContrast}. Likewise, the authors of~\cite{2D-LSTM} used four different LSTM cells, each receiving distinct parts of the image. Recurrent Convolutional Neural Networks (rCNN)~\cite{recurCNN} recurrently train the network using different input window sizes fed into the RNN.
This approach achieves better segmentation and avoids the loss of resolution encountered with fixed window fitting in RNN methods.

\section{Methodology}

We propose an efficient architecture for Semantic Segmentation making use of the large FOV generated by Atrous Convolutions combined with cascade of convolutions in a ``Waterfall'' configuration. Our WASP architecture provides benefits due to its multiscale representations as well as efficiency in the reduced size of the network.

\begin{figure}[H]
\centering
\includegraphics[width=1\linewidth]{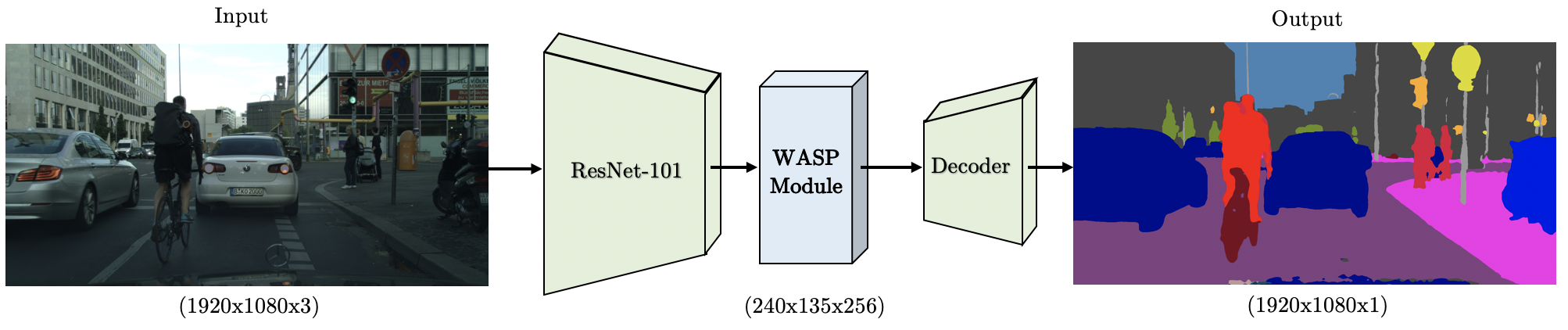}
\caption{WASPnet architecture for semantic segmentation.}
\label{fig:pipeline}
\end{figure}

The processing pipeline is shown in Figure \ref{fig:pipeline}. The input image is initially fed into a deep CNN (namely a ResNet-101 architecture) with the final layers replaced by a WASP module. The resultant score map with the probability distributions obtained from Softmax is processed by a decoder network that performs bilinear interpolation and generates a more efficient segmentation without the use of postprocessing with CRF. We provide a comparison of our WASP architecture with DeepLab's original ASPP architecture and with a modified architecture based on the Res2Net module.

\subsection{Res2Net-Seg Module}
Res2Net~\cite{Res2Net} is a recently developed architecture designed to improve upon  ResNet~\cite{ResNet}.
Res2Net incorporates multiscale features with a Squeeze-and-Excitation (SE) block~\cite{SE} to obtain better representations and achieves promising results. The Res2Net module divides the original bottleneck block into four parallel streams, each containing 25\% of the layers that are fed to 4 different 3 $\times$ 3 convolutions. Simultaneously, it incorporates the output of the parallel convolution.
The SE block is an adaptable architecture that can recalibrate the responses in the feature map channel by modeling the interdependencies between channels. This allows improvements in performance by exploiting the dependencies between feature maps without increase in the network size.

\begin{figure}[H]
\centering
\includegraphics[width=0.5\linewidth]{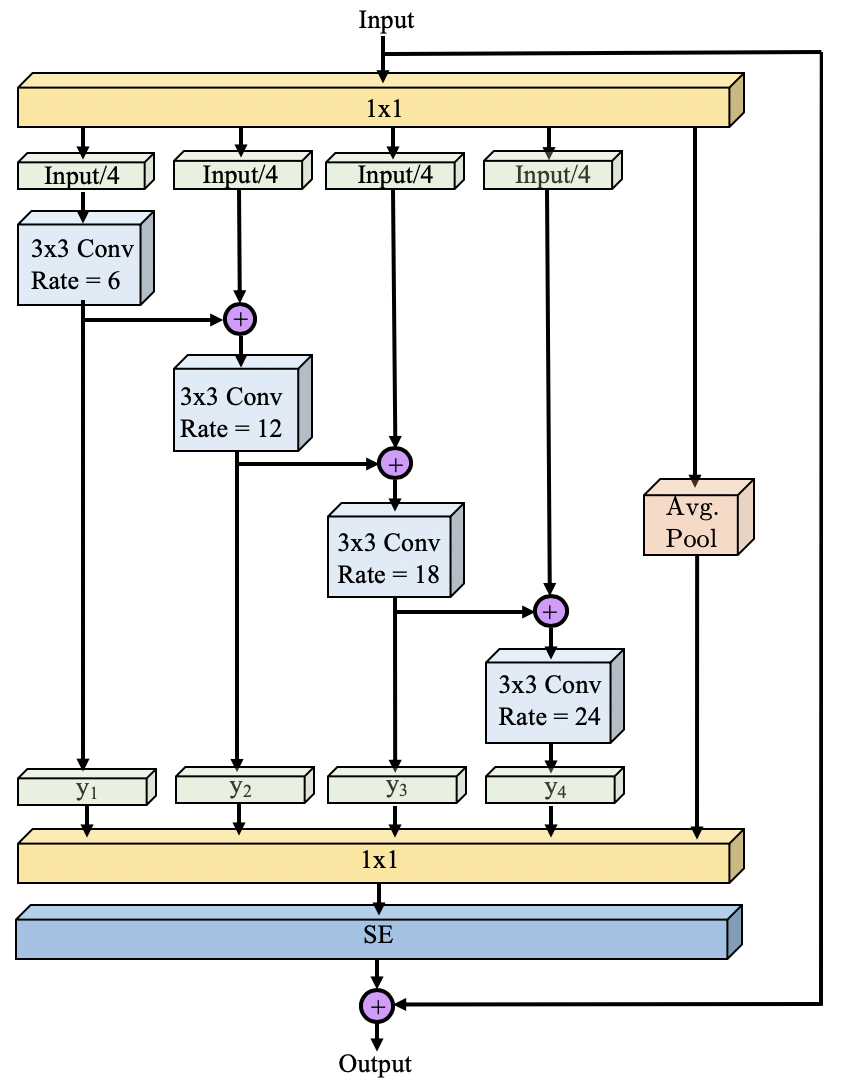}
\caption{Res2Net-Seg block.}
\label{fig:Res2Net}
\end{figure}

Inspired by the work in~\cite{Res2Net}, we present a modified version of the Res2Net module that is suitable for segmentation, named Res2Net-Seg. The Res2Net-Seg module, shown in Figure \ref{fig:Res2Net}, includes the main structure of Res2Net and, additionally, utilizes Atrous Convolutions for each scale for increased FOV and a fifth parallel branch that performs average pooling of all features, which incorporates the original scale in the feature map.
The Res2Net-Seg module is utilized in the WASPnet architecture of Figure \ref{fig:pipeline} in place of the WASP module.
We next propose the WASP module, inspired by multiscale representations, which an improvement over both the Res2Net-Seg and the ASPP configuration.

\subsection{WASP Module}
\begin{figure}[H]
\centering
\includegraphics[width=0.6\linewidth]{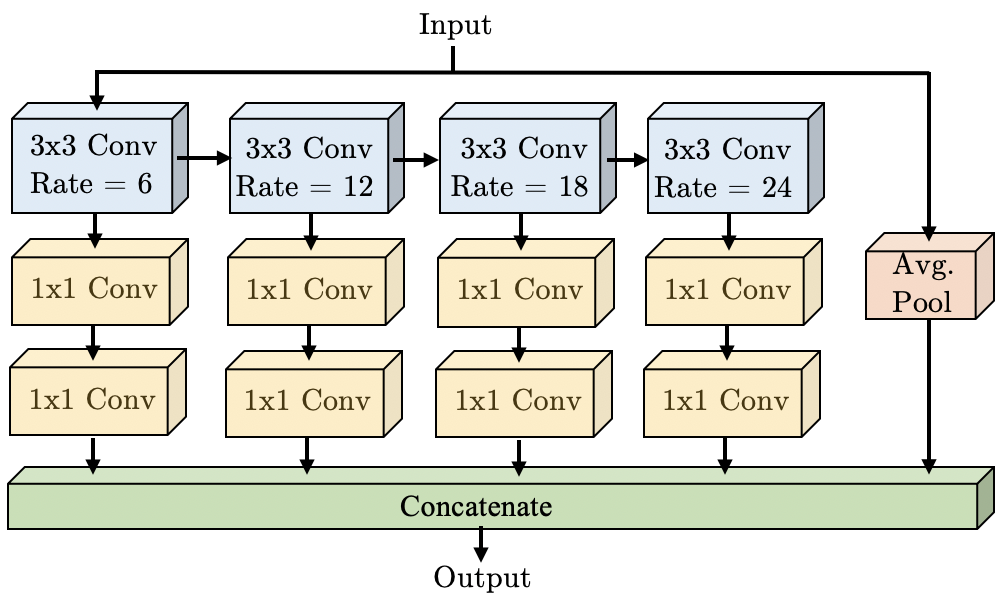}
\caption{Proposed Waterfall Atrous Spatial Pooling (WASP) module.}
\label{fig:WASP}
\end{figure}

We propose the ``Waterfall Atrous Spatial Pyramid'' module, shown in Figure \ref{fig:WASP}.
WASP is a novel architecture with Atrous Convolutions that is able to leverage both the larger FOV of the ASPP configuration and the reduced size of the cascade approach.

An important drawback of Atrous Convolution, applied in either the cascade fashion or the ASPP (parallel design), is that it requires a larger number of parameters and more memory for its implementation, compared to standard convolution.
In~\cite{DeepLab}, there was experimentation to replace convolutional layers of the network backbone architecture, namely, VGG-16 or ResNet-101, with Atrous Convolution modules, but it was too costly in terms of memory requirements. A compromise solution is to apply the cascade of Atrous Convolutions and ASPP modules starting after block 5 when ResNet-101 was utilized.

We overcome these limitations with our Waterfall architecture for improved performance and efficiency.
The Waterfall approach is inspired by multiscale approaches~\cite{Multi-scale-Eigen,Multi-scale-Roy}, the parallel structures of ASPP~\cite{DeepLab}, and Res2Net modules~\cite{Res2Net}, as well as the cascade configuration~\cite{Rethinking}. It is designed with the goal of reducing the number of parameters and memory required, which are the main limitation of Atrous Convolutions.
The WASP module is utilized in the WASPnet architecture shown in Figure \ref{fig:pipeline}.

\begin{figure}[H]
\centering
\includegraphics[width=0.7\linewidth]{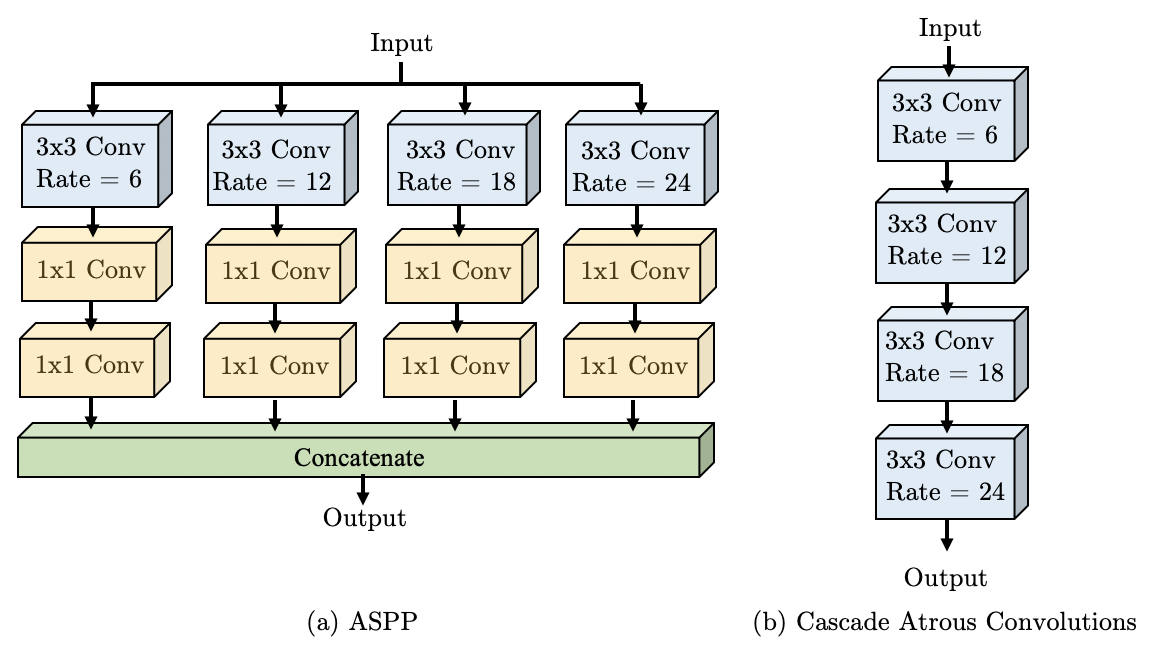}
\caption{Comparison for Atrous Spatial Pyramid Pooling (ASPP)~\cite{DeepLab} and Cascade configuration~\cite{Rethinking}.}
\label{fig:ASPP}
\end{figure}

A comparison between the ASPP module, cascade configuration, and the proposed WASP module is visually highlighted in Figures \ref{fig:WASP} and \ref{fig:ASPP}, for the ASPP and cascade modules.
The WASP configuration consists of four branches of a Large-FOV being fed forward in a waterfall-like fashion. In contrast, the ASPP module uses parallel branches that use more parameter and are less efficient, while the cascade architecture uses sequential filtering operations lacking the larger FOV.

\unskip

\subsection{Decoder}
To process the score maps resulting from the WASP module, a short decoder stage was implemented containing the concatenation with low level features from the first block of the ResNet backbone, convolutional layers, dropout layers, and bilinear interpolations to generate output maps in the same resolution of the input image.

\begin{figure}[H]
\centering
\includegraphics[width=0.9\linewidth]{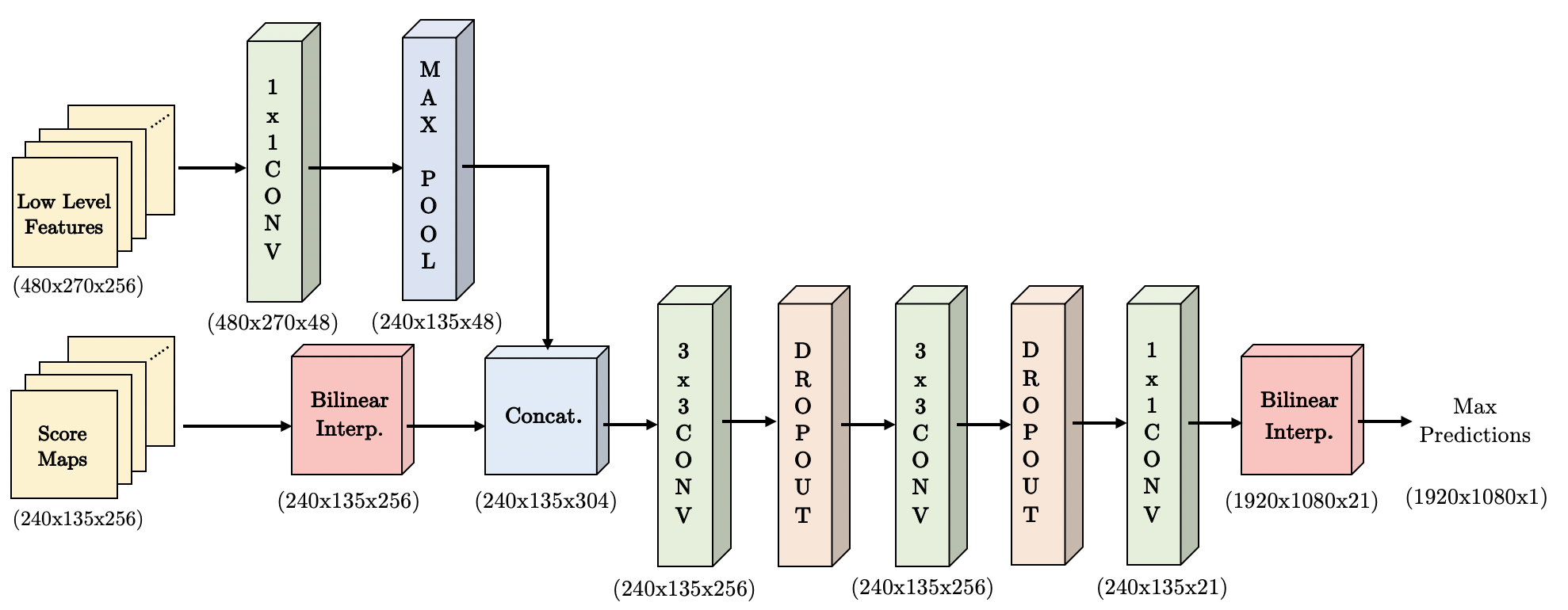}
\caption{Decoder used in the WASPnet method.}
\label{fig:decoder}
\end{figure}

Figure \ref{fig:decoder} shows the decoder and the respective stage dimensions and number of layers. The representation considers an input image with dimensions of 1920 $\times$ 1080 $\times$ 3 for width, height, and RGB color, respectively. In this case, the decoder receives 256 maps of dimensions 240 $\times$ 135 and 256 low level features of dimension 480 $\times$ 270. After matching the dimensions for inputs of the decoder, the layers are concatenated and processed through convolutional layers, dropout, and a final bilinear interpolation to reach the original input size.

\section{Experiments}
\vspace{-6pt}

\subsection{Datasets}
We performed experiments on three datasets used for pre-training, training, validation, and testing.
Microsoft Common Objects in Context (COCO) dataset~\cite{COCO} was used by~\cite{DeepLab} as pre-training as it includes a large amount of data, allowing a good balance of starting weights when training with different datasets, and consequently allowing the increase in precision of the segmentation.

Pascal Visual Object Class (VOC) 2012~\cite{Pascal} is a dataset containing objects in different scenarios including people, animals, vehicles, and indoor objects. It contains three different types of challenges: classification, detection, and segmentation; the latter was utilized in this paper. For the segmentation benchmark, the dataset contains 1464 images for training, 1449 images for validation, and 1456 images for testing annotated for 21 classes. Data augmentation was used to increase the training set size to~10,582.

Cityscapes~\cite{Cityscapes} is a larger dataset containing urban scene images recorded in street scenes of 50~different cities with pixel annotations of 25,000 frames.
In the Cityscapes dataset, 5000 images are finely annotated at pixel level divided into 2975 images for training, 500 for validation, and 1525 for testing. Cityscapes is annotated in 19 semantic classes divided into 7 categories (construction, ground, human, nature, object, sky, and vehicle).

\subsection{Evaluation Metrics}
We based our comparison of performance to other methods using Mean Intersection over Union (mIOU), considered the most important and more widely used metric for semantic segmentation. A~pixel-level analysis of detection is conducted, reporting the intersection of true positive (TP) pixels detection as a percentage of the union of TP with false negative (FN) and false positive (FP) pixels.

\subsection{Simulation Parameters}
We calculate the learning rate based on the polynomial method (``poly'')~\cite{Parsenet}, also adopted in~\cite{DeepLab}. The poly learning rate $LR_{poly}$ results in more effective updating of the weights when compared to the traditional ``step'' learning rate, given as
\begin{equation}
LR_{poly}=(1-\frac{iter}{max\_iter})^{power}
\end{equation}
where $power\!=\!0.9$ was employed.
We utilized a batch size of eight due to physical memory constraints in the hardware available, lower than the batch size of ten used by DeepLab. A subtle improvement in training with a larger batch size is expected for the architectures proposed.

We experimented with different rates of dilation on WASP. We found that larger rates result in better mIOU. A set rate of $r =$ \{6, 12, 18, 24\} was selected for the WASP module. In addition, we performed pre-training using the MS-COCO dataset~\cite{COCO}, and data augmentation in randomly selected images scaled between (0.5,1.5).

\section{Results}

Following training, validation, and testing procedures, the WASPnet architecture was implemented utilizing WASP module, Res2Net-Seg module, or ASPP module.
The validation mIOU results are presented in Table \ref{tab:TrainingVOC} for the Pascal VOC dataset.
When following similar guidelines as in~\cite{DeepLab} for training and hyperparameters, and using the WASP module, an mIOU of 80.22\% is achieved without the need for CRF postprocessing. Our WASPnet resulted in a gain of 5.07\% on the validation set and reduced the number of parameters by 20.69\%.

\begin{table}[H]
\caption{Pascal Pascal Visual Object Class (VOC) validation set results.}
\centering
\begin{tabular}{cccc}
\toprule
\textbf{Architecture}& \textbf{Number of Parameters} & \textbf{Parameter Reduction} & \textbf{mIOU}\\
\midrule
WASPnet-CRF (ours)&47.482 M&20.69\% &80.41\%\\
WASPnet (ours)&47.482 M&20.69\%&80.22\%\\
Res2Net-Seg-CRF&50.896 M&14.99\%&80.12\%\\
Res2Net-Seg&50.896 M&14.99\%&78.53\%\\
Deeplab-CRF~\cite{DeepLab}&59.869 M&-&77.69\%\\
Deeplab~\cite{DeepLab}&59.869 M&-&76.35\%\\
\bottomrule
\end{tabular}
\label{tab:TrainingVOC}
\end{table}

The Res2Net-Seg approach results in an mIOU of 78.53\% without CRF, achieves  mIOU  of 80.12\% with CRF, and reduces the number of parameters by 14.99\%. The Res2Net-Seg approach still shows benefits with the incorporation of CRF as postprocessing, similar to the cascade and ASPP methods.

\begin{figure}[H]
\centering
\includegraphics{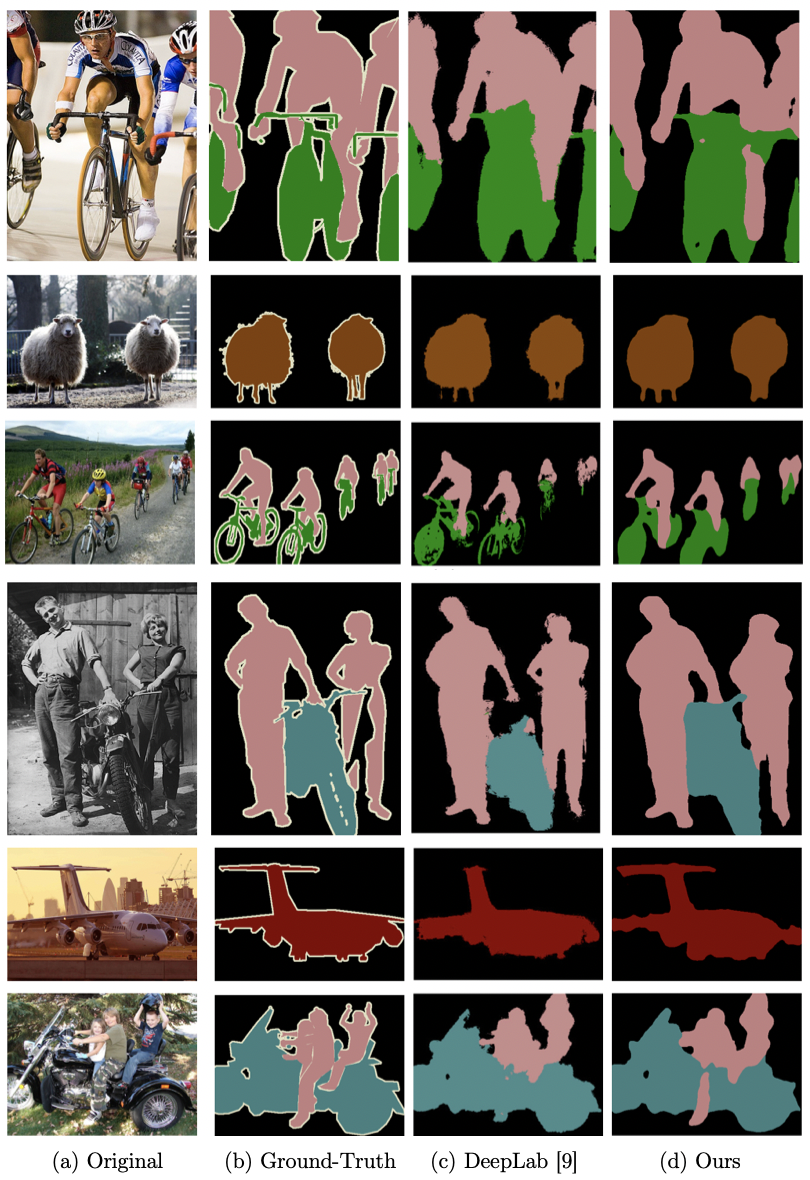}
\caption{Results sample for Pascal VOC dataset~\cite{Pascal}.}
\label{fig:ImagesVOC}
\end{figure}

Overall, the WASP architecture provides the best result and the highest reduction in parameters.
Sample results for the WASPnet architecture are shown in Figure \ref{fig:ImagesVOC} for validation images from the Pascal VOC dataset~\cite{Pascal}. Note, from the generated segmentation, that our method presents a better definition in the detection shape, being closer to the ground-truth when compared to previous methods utilizing ASPP (DeepLab).

We tested the effects of different dilation rates (in our WASP module) on the final segmentation. In our tests, all kernel sizes were set to 3 following procedures as in~\cite{DeepLab}. Table \ref{tab:DilationRates} reports the accuracy, in mIOU, for the Pascal VOC dataset for different dilation rates in the WASP module.
The configuration with dilation rates of \{6, 12, 18, 24\} resulted in the best accuracy for the Pascal VOC dataset, therefore, the following tests were conducted using this dilation rate.

\begin{table}[H]
\caption{Pascal VOC validation set results for different sets of dilation in the WASP module.}
\centering
\begin{tabular}{cc}
\toprule
\textbf{WASP Dilation Rates} & \textbf{mIOU}\\
\midrule
\{2, 4, 6, 8\}&79.61\%\\
\{4, 8, 12, 16\}&79.72\%\\
\{6, 12, 18, 24\}&80.22\%\\
\{8, 16, 24, 32\}&79.92\%\\
\bottomrule
\end{tabular}
\label{tab:DilationRates}
\end{table}

We also experimented with postprocessing using CRF. The application of CRF has the benefit of better defining the shapes of the segmented areas.
Similarly to the procedures followed in~\cite{DeepLab}, we performed parameter tuning, for the parameters of Equation (3), by varying $\omega_{1}$ between 3 and 6, $\sigma_{\alpha}$ from 30 to 100, and $\sigma_{\beta}$ from 3 to 6, while fixing both $\omega_{2}$ and $\sigma_{\gamma}$ to 3.

The addition of CRF postprocessing to our WASPnet method resulted in a modest increase of 0.2\% in the mIOU for both the validation and test sets of the Pascal VOC dataset. The gains from using CRF are less significant than those in~\cite{DeepLab}, due to more efficient use of FOV by WASPnet.
The effects of CRF on accuracy were not consistent across different classes.
Classes with objects that do not have extremities, such as bottle, car, bus, and train, benefited most, whereas there was a decrease in accuracy for classes with more delicate boundaries such as bicycle, plant, and motorcycle.

Results on the testing Pascal VOC dataset are shown in Table \ref{tab:TestingVOC}.
The additional training dataset column refers to DeepLabv3 types of models where a ResNet-101 model was pretrained on both ImageNet~\cite{ImageNet} and JFT-300M~\cite{Revisiting} when performing the test challenge for Pascal VOC.
JFT-300M consists of Google's internal dataset of 300 million images labeled in 18,291 categories, and therefore these results cannot be compared directly to other external architectures including this work. The addition of the JFT dataset for training allows the architecture to achieve performance improvements that are not possible without the such a large number of training samples.
Note that training of the WASPnet network was performed only on the training dataset provided by the challenge, consisting of 1464 images.
Based on these results, WASPnet outperforms all of the other methods that are trained on the same dataset.

\begin{table}[H]
\caption{Pascal VOC test set results.}
\centering
\begin{tabular}{ccc}
\toprule
\textbf{Architecture}&\textbf{Additional Training Dataset Used}&\textbf{mIOU}\\

\midrule
DeepLabv3+~\cite{DeepLabv3+}&JFT-300M~\cite{Revisiting}&87.8\%\\
Deeplabv3~\cite{Rethinking}&JFT-300M~\cite{Revisiting}&85.7\%\\
Auto-DeepLab-L~\cite{Auto-DeepLab}&JFT-300M~\cite{Revisiting}&85.6\%\\
Deeplab~\cite{DeepLab}&JFT-300M~\cite{Revisiting}&79.7\%\\
\midrule
WASPnet-CRF (ours)&-&79.6\%\\
WASPnet (ours)&-&79.4\%\\
Dilation~\cite{DilatedConv}&-&75.3\%\\
CRFasRNN~\cite{CRFasRNN}&-&74.7\%\\
ParseNet~\cite{Parsenet}&-&69.8\%\\
FCN 8s~\cite{FCN}&-&67.2\%\\
Bayesian SegNet~\cite{Bayesian-SegNet}&-&60.5\%\\
\bottomrule
\end{tabular}
\label{tab:TestingVOC}
\end{table}

WASPnet was also used with the Cityscapes dataset~\cite{Cityscapes} following similar procedures. Table \ref{tab:TrainingCityscapes} shows the results obtained for Cityscapes, resulting in an mIOU of 74.0\%, a gain of 4.2\% from~\cite{DeepLab}.
The Res2Net-Seg version of the network achieved 72.1\% mIOU.

\begin{table}[H]
\caption{Cityscapes validation set results.}
\centering
\begin{tabular}{cccc}
\toprule
\textbf{Architecture}& \textbf{Number of Parameters} & \textbf{Parameter Reduction} & \textbf{mIOU}\\
\midrule
WASPnet (ours)&47.482 M&20.69\%&74.0\%\\
WASPnet-CRF (ours)&47.482 M&20.69\%&73.2\%\\
Res2Net-Seg (ours)&50.896 M&14.99\%&72.1\%\\
Deeplab-CRF~\cite{DeepLab}&59.869 M&-&71.4\%\\
Deeplab~\cite{DeepLab}&59.869 M&-&71.0\%\\
\bottomrule
\end{tabular}
\label{tab:TrainingCityscapes}
\end{table}

For both WASP and Res2Net-Seg architectures tested on the Cityscapes dataset, the CRF postprocessing did not have much benefit. A  similar result was found with DeepLab where CRF resulted in a small improvement of the mIOU.
The higher resolution and shape of detected instances in the Cityscapes dataset likely affected the effectiveness of the CRF.
With Cityscapes, we used a batch size of 4 due to hardware constraints during training; other architectures have used batch sizes of up to ten.

Table \ref{tab:TestingCityscapes} shows the results of WASPnet on the Cityscapes testing dataset. WASPnet achieved mIOU of 70.5\% and outperformed other architectures trained on the dame dataset.
We only performed training on the fine annotation images from the Cityscapes dataset, containing 2975 images, whereas the DeepLabv3 style architectures used larger datasets for training, such as JFT-300M containing 300~million images for pre-training and and coarser dataset from Cityscapes containing 20,000 images.

\begin{table}[H]
\caption{Pascal Cityscapes test set results.}
\centering
\begin{tabular}{ccc}
\toprule
\textbf{Architecture}&\textbf{Additional Training Dataset Used}&\textbf{mIOU}\\
\midrule
Auto-DeepLab-L~\cite{Auto-DeepLab}&Coarse Cityscapes~\cite{Cityscapes}&82.1\%\\
DeepLabv3+~\cite{DeepLabv3+}&Coarse Cityscapes~\cite{Cityscapes}&82.1\%\\
\midrule
WASPnet (ours)&-&70.5\%\\
Deeplab~\cite{DeepLab}&-&70.4\%\\
Dilation~\cite{DilatedConv}&-&67.1\%\\
FCN 8s~\cite{FCN}&-&65.3\%\\
CRFasRNN~\cite{CRFasRNN}&-&62.5\%\\
ENet~\cite{Enet}&-&58.3\%\\
SegNet~\cite{Segnet}&-&55.6\%\\
Mask-RCNN~\cite{MaskRCNN}&-&49.9\%\\
\bottomrule
\end{tabular}
\label{tab:TestingCityscapes}
\end{table}

\begin{figure}[H]
\centering
\includegraphics[width=0.7\linewidth]{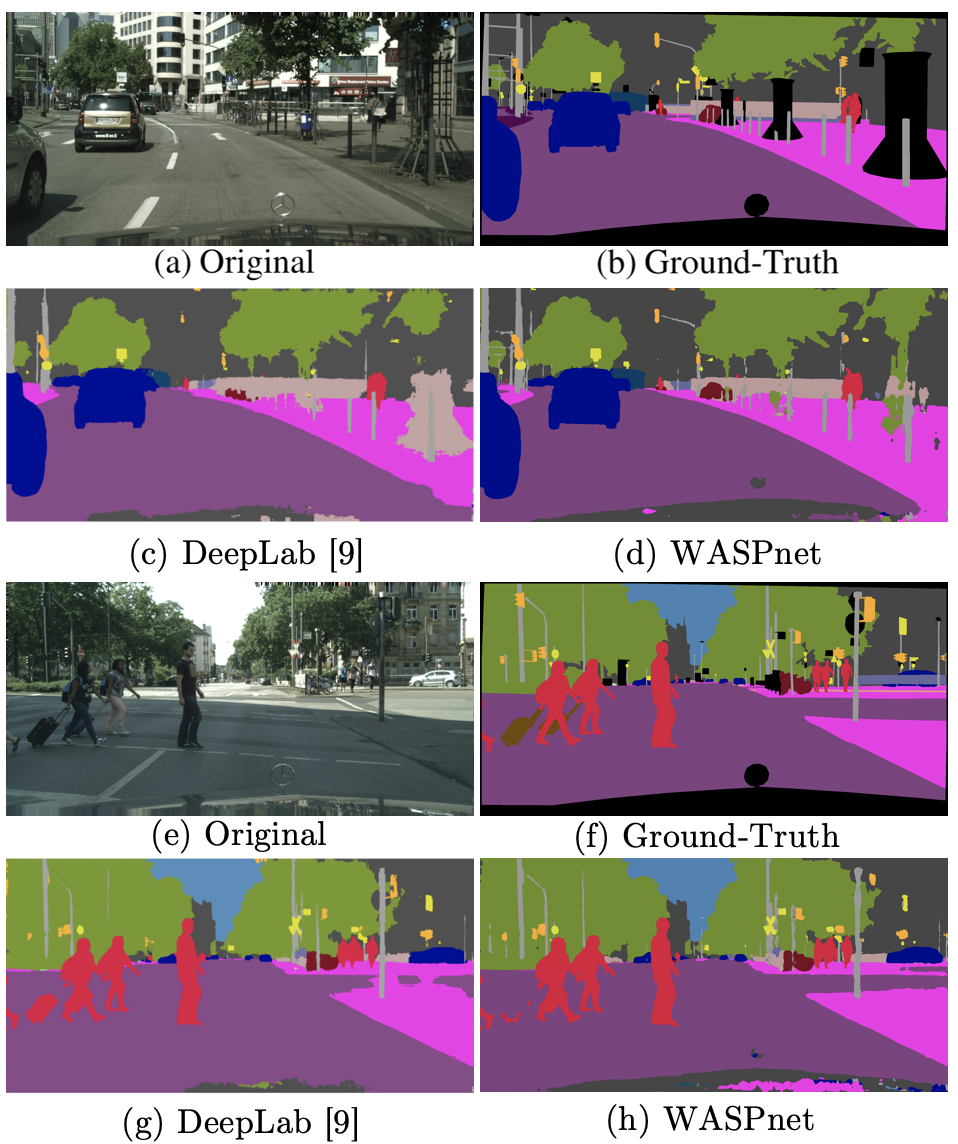}
\caption{Results sample for Cityscapes dataset~\cite{Cityscapes}.}
\label{fig:cityscapes}
\end{figure}

Figure \ref{fig:cityscapes} shows examples of Cityscapes image segmentations with the WASPnet method. Like our observations from the Pascal VOC dataset, our method produces better defined shapes for the segmentation compared to DeepLab.
Our results are closer to the ground-truth data, and show better segmentation of smaller objects that are further away from the camera.

Our results in Table \ref{tab:TrainingCityscapes} illustrate that  postprocessing with CRF slightly decreased the mIOU by 0.8\% in the Cityscapes dataset: CRF has difficulty dealing with delicate boundaries, which are common in the Cityscapes dataset.
With WASPnet, the presence of larger FOV due to the WASP module is able to offset the potential gains of the CRF module from previous networks.
An additional limitation is that CRF requires substantial extra time for processing.  For these reasons, we conclude that WASPnet can be used without CRF postprocessing.

\subsection*{Fail Cases}

\begin{figure}[H]
\centering
\includegraphics[width=1\linewidth]{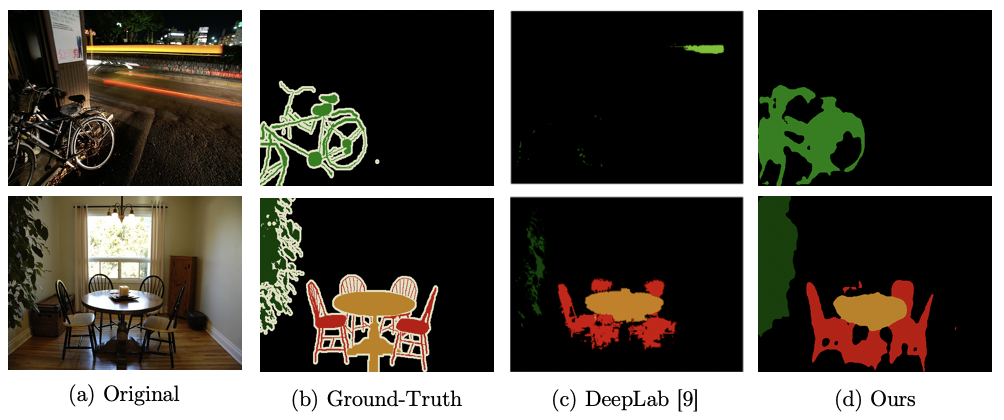}
\caption{Occurrence of fail cases to detect more delicate boundaries}
\label{fig:Fail}
\end{figure}

Classes that contain more delicate, and consequently harder to accurately detect, shapes contribute the most to segmentation errors. Particularly, tables, chairs, leaves, and bicycles present a bigger challenge to segmentation networks.
These classes also resulted in a lower accuracy when applying CRF.
Representative examples of fail cases are shown in Figure \ref{fig:Fail} for classes chair and bicycle, which are the most difficult to segment. Even in these cases, WASPnet (without CRF) is able to better detect the general shape compared to DeepLab.

\section{Conclusions}
We propose a ``Waterfall'' architecture based on the WASP module for efficient semantic segmentation that achieves high mIOU scores on the Pascal VOC and Cityscapes datasets. The smaller size of this efficient architecture improves its functionality and reduces the risk of overfitting without the need for postprocessing with the time consuming CRF.
The results of WASPnet segmentation demonstrated superior performance compared to Res2Net-Seg and Deeplab.
This work provides the foundation for further application of WASP in a broader range of applications for more efficient multiscale analysis.

\vspace{6pt}

\authorcontributions{
Conceptualization, B.A. and A.S.; methodology, B.A.; algorithm and experiments, B.A. and A.S.; original draft preparation, B.A. and A.S.; review and editing, B.A. and A.S.; supervision, A.S.; project administration, A.S.; funding acquisition, A.S.}


\funding{This research was funded in part by National Science Foundation grant number 1749376.}


\conflictsofinterest{The authors declare no conflict of interest.}

\abbreviations{The following abbreviations are used in this manuscript:\\

\noindent
\begin{tabular}{@{}ll}
ASPP & Atrous Spatial Pyramid Pooling\\
COCO & Common Objects in Context\\
CNN & Convolutional Neural Networks\\
CRF & Conditional Random Fields\\
ENet & Efficient Neural Network\\
FCN & Fully Convolutional Networks\\

\end{tabular}

\noindent
\begin{tabular}{@{}ll}
FN & False Negative\\
FOV & Field-of-View\\
FP & False Positive\\
LSTM & Long Short-Term Memory\\
LSTM-CF & Long Short-Term Memory Context Fusion\\
rCNN & Recurrent Convolutional Neural Networks\\
mIOU & Mean Intersection over Union\\
RGB & Red, Green, and Blue\\
RNN & Recurrent Neural Networks\\
SE & Squeeze-and-Excitation\\
TP & True Positive\\
VOC & Visual Object Class\\
WASP & Waterfall Atrous Spatial Pooling\\
\end{tabular}}

\reftitle{References}


\begin{thebibliography}{999}
\providecommand{\natexlab}[1]{#1}

\bibitem[Garcia{-}Garcia \em{et~al.}(2017)Garcia{-}Garcia, Orts{-}Escolano,
Oprea, Villena{-}Martinez, and Rodr{\'{\i}}guez]{Review}
Garcia{-}Garcia, A.; Orts{-}Escolano, S.; Oprea, S.; Villena{-}Martinez, V.;
Rodr{\'{\i}}guez, J.G.
\newblock A Review on Deep Learning Techniques Applied to Semantic
Segmentation.
\newblock {\em arXiv} {\bf 2017}, arXiv:1704.06857.

\bibitem[Zhu \em{et~al.}(2016)Zhu, Meng, Cai, and Lu]{Review2}
Zhu, H.; Meng, F.; Cai, J.; Lu, S.
\newblock A comprehensive survey from bottom-up to semantic image segmentation
and cosegmentation.
\newblock {\em J. Vis. Commun. Image Represent.} {\bf
2016}, {\em 34},~12--27. [\href{http://dx.doi.org/10.1016/j.jvcir.2015.10.012}{CrossRef}]

\bibitem[Thoma(2016)]{Review3}
Thoma, M.
\newblock A Survey of Semantic Segmentation.
\newblock {\em arXiv} {\bf 2016}, arXiv:1602.0654.

\bibitem[Ess \em{et~al.}(2009)Ess, M\"{u}ller, Grabner, and Goo]{Driving}
Ess, A.; M\"{u}ller, T.; Grabner, H.; Goo, L.J.V.
\newblock Segmentation-based urban traffic scene understanding.
\newblock {\em BMVC} {\bf 2009}, {\em 1},~2.

\bibitem[Oberweger and Vincent Lepetit(2015)]{HMI}
Oberweger, M.; Wohlhart, P.; Lepetit, V.
\newblock Hands Deep in Deep Learning for Hand Pose Estimation.
\newblock {\em arXiv} {\bf 2015}, arXiv:1502.06807.

\bibitem[Fan \em{et~al.}(2017)Fan, Liu, Xiong, and Wu]{photo}
Fan, H.; Liu, D.; Xiong, Z.; Wu, F.
\newblock Two-stage convolutional neural network for light field
super-resolution.
\newblock In Proceedings of the Image Processing (ICIP) 2017 IEEE International Conference, Beijing, China, 17--20 September
2017; pp. 1167--1171.

\bibitem[Tzelepi and Tefas(2018)]{ImageSearch}
Tzelepi, M.; Tefas, A.
\newblock Deep convolutional learning for content based image retrieval.
\newblock {\em Neurocomputing} {\bf 2018}, {\em 275},~2467--2478. [\href{http://dx.doi.org/10.1016/j.neucom.2017.11.022}{CrossRef}]

\bibitem[Long \em{et~al.}(2015)Long, Shelhamer, and Darrel]{FCN}
Long, J.; Shelhamer, E.; Darrel, T.
\newblock Fully Convolutional Networks for Semantic Segmentation.
\newblock In Proceedings of the  2015 IEEE Conference on Computer Vision and Pattern Recognition (CVPR), Boston, MA, USA, \mbox{7--12 June 2015.}

\bibitem[Chen \em{et~al.}(2018)Chen, Papandreou, Kokkinos, Murphy, and
Yuille]{DeepLab}
Chen, L.C.; Papandreou, G.; Kokkinos, I.; Murphy, K.; Yuille, L.
\newblock DeepLab: Semantic Image Segmentation with Deep Convolutional Nets,
Atrous Convolution and Fully Connected CFRs.
\newblock {\em IEEE Trans. Pattern Anal. Mach. Intell.}
{\bf 2018}, {\em 40},~834--845. [\href{http://dx.doi.org/10.1109/TPAMI.2017.2699184}{CrossRef}]

\bibitem[Chen \em{et~al.}(2017)Chen, Papandreou, Schroff, and Adam]{Rethinking}
Chen, L.C.; Papandreou, G.; Schroff, F.; Adam, H.
\newblock Rethinking Atrous Convolution for Semantic Image Segmentation.
\newblock {\em arXiv} {\bf 2017}, arXiv:1706.05587.

\bibitem[Gao \em{et~al.}(2020)Gao, Cheng, Zhao, Zhang, Yang, and Torr]{Res2Net}
Gao, S.H.; Cheng, M.M.; Zhao, K.; Zhang, X.Y.; Yang, M.H.; Torr, P.
\newblock Res2Net: A New Multi-Scale Backbone Architecture.
\newblock {\em IEEE Trans. Pattern Anal. Mach. Intell.}
{\bf 2019}. [\href{http://dx.doi.org/10.1109/TPAMI.2019.2938758}{CrossRef}]

\bibitem[Krizhevsky \em{et~al.}(2012)Krizhevsky, Sutskever, and
Hinton]{AlexNet}
Krizhevsky, A.; Sutskever, I.; Hinton, G.E.
\newblock ImageNet Classification with Deep Convolutional Neural Networks.
\newblock In Proceedings of the  Advances in Neural Information Processing Systems 25 (NIPS), Lake Tahoe, NV, USA, 3--6 December 2012.   

\bibitem[Simonyan and Zisserman(2015)]{VGG}
Simonyan, K.; Zisserman, A.
\newblock Very Deep Convolutional Networks for Large-Scale Image Recognition.
\newblock {\em arXiv} {\bf 2015}, arXiv:1409.1556.

\bibitem[Szegedy \em{et~al.}(2014)Szegedy, Liu, Jia, Sermanet, Reed, Anguelov,
Erhan, Vanhoucke, and Rabinovich]{GoogleNet}
Szegedy, C.; Liu, W.; Jia, Y.; Sermanet, P.; Reed, S.E.; Anguelov, D.; Erhan,
D.; Vanhoucke, V.; Rabinovich, A.
\newblock Going Deeper with Convolutions.
\newblock {\em arXiv} {\bf 2014}, arXiv:1409.4842.

\bibitem[He \em{et~al.}(2015)He, Zhang, Ren, and Sun]{ResNet}
He, K.; Zhang, X.; Ren, S.; Sun, J.
\newblock Deep Residual Learning for Image Recognition.
\newblock {\em arXiv} {\bf 2015}, arXiv:1512.03385.

\bibitem[Yu and Koltun(2016)]{DilatedConv}
Yu, F.; Koltun, V.
\newblock Multi-Scale Context Aggregation by Dilated Convolutions.
\newblock In Proceedings of the  ICLR, San Juan, PR, USA, 2--4 May 2016.

\bibitem[Chen \em{et~al.}(2018)Chen, Zhu, Papandreou, Schroff, and
Adam]{DeepLabv3+}
Chen, L.; Zhu, Y.; Papandreou, G.; Schroff, F.; Adam, H.
\newblock Encoder-Decoder with Atrous Separable Convolution for Semantic Image
Segmentation.
\newblock {\em arXiv} {\bf 2018}, arXiv:1802.02611.

\bibitem[Paszke \em{et~al.}(2016)Paszke, Chaurasia, Kim, and Culurciello]{Enet}
Paszke, A.; Chaurasia, A.; Kim, S.; Culurciello, E.
\newblock ENet: {A} Deep Neural Network Architecture for Real-Time Semantic
Segmentation.
\newblock {\em arXiv} {\bf 2016}, arXiv:1606.02147.

\bibitem[He \em{et~al.}(2014)He, Zhang, Ren, and Sun]{SPP}
He, K.; Zhang, X.; Ren, S.; Sun, J.
\newblock Spatial Pyramid Pooling in Deep Convolutional Networks for Visual
Recognition.
\newblock {\em arXiv} {\bf 2014}, arXiv:1406.4729.

\bibitem[Dai \em{et~al.}(2017)Dai, Qi, Xiong, Li, Zhang, Hu, and
Wei]{Deformable}
Dai, J.; Qi, H.; Xiong, Y.; Li, Y.; Zhang, G.; Hu, H.; Wei, Y. Deformable convolutional networks.
\newblock  In Proceedings of the IEEE International Conference on Computer Vision, Venice, Italy, 22--29 October 2017; pp. 764--773.  


\bibitem[Deng \em{et~al.}(2009)Deng, Dong, Socher, Li, Li, and
Fei-Fei]{ImageNet}
Deng, J.; Dong, W.; Socher, R.; Li, L.J.; Li, K.; Fei-Fei, L.
\newblock {ImageNet: A Large-Scale Hierarchical Image Database}.
\newblock  In Proceedings of the Conference of Computer Vision and Pattern Recognition (CVPR), Miami, FL, USA,  20--25 June  2009.

\bibitem[Sun \em{et~al.}(2017)Sun, Shrivastava, Singh, and Gupta]{Revisiting}
Sun, C.; Shrivastava, A.; Singh, S.; Gupta, A.
\newblock Revisiting Unreasonable Effectiveness of Data in Deep Learning Era.
\newblock {\em arXiv} {\bf 2017}, arXiv:1707.02968.

\bibitem[Badrinarayanan \em{et~al.}(2015)Badrinarayanan, Kendall, and
Cipolla]{Segnet}
Badrinarayanan, V.; Kendall, A.; Cipolla, R.
\newblock SegNet: {A} Deep Convolutional Encoder-Decoder Architecture for Image
Segmentation.
\newblock {\em arXiv} {\bf 2015}, arXiv:1511.00561.

\bibitem[Liu \em{et~al.}(2019)Liu, Chen, Schroff, Adam, Hua, Yuille, and
Fei-Fei]{Auto-DeepLab}
Liu, C.; Chen, L.C.; Schroff, F.; Adam, H.; Hua, W.; Yuille, A.L.; Fei-Fei, L.
\newblock Auto-DeepLab: Hierarchical Neural Architecture Search for Semantic
Image Segmentationx.
\newblock  In Proceedings of the  IEEE Conference on Computer Vision and Pattern Recognition
(CVPR), Long Beach, CA, USA, 16--20 June 2019; pp. 82--92. 

\bibitem[Zheng \em{et~al.}(2015)Zheng, Jayasumana, Romera{-}Paredes, Vineet,
Su, Du, Huang, and Torr]{CRFasRNN}
Zheng, S.; Jayasumana, S.; Romera{-}Paredes, B.; Vineet, V.; Su, Z.; Du, D.;
Huang, C.; Torr, P.H.S.
\newblock Conditional Random Fields as Recurrent Neural Networks.
\newblock {\em arXiv} {\bf 2015}, arXiv:1502.03240.

\bibitem[Kr\"{a}hen\"{u}hl and Koltun(2011)]{CRF}
Kr\"{a}hen\"{u}hl, P.; Koltun, V.
\newblock Efficient Inference in Fully Connected CRFs with Gaussian Edge
Potentials.
\newblock  In~Proceedings of the NIPS, Granada, Spain, 12--17 December 2011. 

\bibitem[Raj \em{et~al.}(2015)Raj, Maturana, and Scherer]{Multi-scale-Raj}
Raj, A.; Maturana, D.; Scherer, S.
\newblock \emph{Multi-Scale Convolutional Architecture for Semantic Segmentation};
\newblock  Robotics Institute, Carnegie Mellon University, Tech.:
Pittsburgh, PA, USA, 2015.  

\bibitem[Eigen and Fergus(2014)]{Multi-scale-Eigen}
Eigen, D.; Fergus, R.
\newblock Predicting Depth, Surface Normals and Semantic Labels with a Common
Multi-Scale Convolutional Architecture.
\newblock {\em arXiv} {\bf 2014}, arXiv:1411.4734.

\bibitem[Roy and Todorovic(2016)]{Multi-scale-Roy}
Roy, A.; Todorovic, S.
\newblock A Multi-Scale CNN for Affordance Segmentation in RGB Images.
\newblock In~Proceedings of the IEEE European Conference on Computer
Vision (ECCV), Amsterdam, the Netherlands, \mbox{11--14 October 2016}; pp. 186--201. 

\bibitem[Bian \em{et~al.}(2016)Bian, Lim, and Zhou]{Multi-scale-Bian}
Bian, X.; Lim, S.N.; Zhou, N.
\newblock Multiscale fully convolutional network with application to industrial
inspection.
\newblock In Proceedings of the  IEEE Winter Conference on Applications of Computer Vision
(WACV),  Lake Placid, NY, USA,  7--10 March 2016; pp. 1--8.

\bibitem[Kendall \em{et~al.}(2015)Kendall, Badrinarayanan, and
Cipolla]{Bayesian-SegNet}
Kendall, A.; Badrinarayanan, V.; Cipolla, R.
\newblock Bayesian SegNet: Model Uncertainty in Deep Convolutional
Encoder-Decoder Architectures for Scene Understanding.
\newblock {\em arXiv} {\bf 2015}, arXiv:1511.02680.

\bibitem[Liu \em{et~al.}(2015)Liu, Rabinovich, and Berg]{Parsenet}
Liu, W.; Rabinovich, A.; Berg, A.C.
\newblock ParseNet: Looking Wider to See Better.
\newblock {\em arXiv} {\bf 2015}, arXiv:1506.04579.

\bibitem[Visin \em{et~al.}(2015{\natexlab{a}})Visin, Kastner, Courville,
Bengio, Matteucci, and Cho]{ReSeg}
Visin, F.; Kastner, K.; Courville, A.C.; Bengio, Y.; Matteucci, M.; Cho, K.
\newblock ReSeg: {A} Recurrent Neural Network for Object Segmentation.
\newblock {\em arXiv} {\bf 2015}, arXiv:1511.07053.

\bibitem[Visin \em{et~al.}(2015{\natexlab{b}})Visin, Kastner, Cho, Matteucci,
Courville, and Bengio]{ReNet}
Visin, F.; Kastner, K.; Cho, K.; Matteucci, M.; Courville, A.C.; Bengio, Y.
\newblock ReNet: {A} Recurrent Neural Network Based Alternative to
Convolutional Networks.
\newblock {\em arXiv} {\bf 2015}, arXiv:1505.00393.

\bibitem[Li \em{et~al.}(2016)Li, Gan, Liang, Yu, Cheng, and Lin]{LSTM-CF}
Li, Z.; Gan, Y.; Liang, X.; Yu, Y.; Cheng, H.; Lin, L.
\newblock {RGB-D} Scene Labeling with Long Short-Term Memorized Fusion Model.
\newblock {\em arXiv} {\bf 2016}, arXiv:1604.05000.

\bibitem[Li and Yu(2016)]{DeepContrast}
Li, G.; Yu, Y.
\newblock Deep Contrast Learning for Salient Object Detection.
\newblock {\em arXiv} {\bf 2016}, arXiv:1603.01976.

\bibitem[Byeon \em{et~al.}(2015)Byeon, Breuel, Raue, and Liwicki]{2D-LSTM}
Byeon, W.; Breuel, T.M.; Raue, F.; Liwicki, M.
\newblock Scene labelingwith lstm recurrent neural networks.
\newblock In~Proceedings of the IEEE Conference on Computer Vision and Pattern
Recognition,  Boston, MA, USA,  7--12 June  2015; pp. 3547--3555.

\bibitem[Pinheiro and Collobert(2013)]{recurCNN}
Pinheiro, P.H.O.; Collobert, R.
\newblock Recurrent Convolutional Neural Networks for Scene Parsing.
\newblock {\em arXiv} {\bf 2013}, arXiv:1306.2795.

\bibitem[Hu \em{et~al.}(2017)Hu, Shen, and Sun]{SE}
Hu, J.; Shen, L.; Sun, G.
\newblock Squeeze-and-Excitation Networks.
\newblock {\em arXiv} {\bf 2017}, arXiv:1709.01507.

\bibitem[Lin \em{et~al.}(2014)Lin, Maire, Belongie, Bourdev, Girshick, Hays,
Perona, Ramanan, Doll{\'{a}}r, and Zitnick]{COCO}
Lin, T.; Maire, M.; Belongie, S.J.; Bourdev, L.D.; Girshick, R.B.; Hays, J.;
Perona, P.; Ramanan, D.; Doll{\'{a}}r, P.; Zitnick, C.L.
\newblock Microsoft {COCO:} Common Objects in Context.
\newblock {\em arXiv} {\bf 2014}, arXiv:1405.0312.

\bibitem[Everingham \em{et~al.}(2010)Everingham, Van~Gool, Williams, Winn, and
Zisserman]{Pascal}
Everingham, M.; Van~Gool, L.; Williams, C.K.I.; Winn, J.; Zisserman, A.
\newblock The Pascal Visual Object Classes (VOC) Challenge.
\newblock {\em Int. J. Comput. Vis.} {\bf 2010}, {\em
88},~303--338. [\href{http://dx.doi.org/10.1007/s11263-009-0275-4}{CrossRef}]

\newpage

\bibitem[Cordts \em{et~al.}(2016)Cordts, Omran, Ramos, Rehfeld, Enzweiler,
Benenson, Franke, Roth, and Schiele]{Cityscapes}
Cordts, M.; Omran, M.; Ramos, S.; Rehfeld, T.; Enzweiler, M.; Benenson, R.;
Franke, U.; Roth, S.; Schiele, B.
\newblock The Cityscapes Dataset for Semantic Urban Scene Understanding.
\newblock  In Proceedings of the  IEEE Conference on Computer Vision and Pattern
Recognition (CVPR),  Las Vegas, NV, USA, 26 June--1 July 2016; pp. 3213--3223. 

\bibitem[He \em{et~al.}(2017)He, Gkioxari, Doll{\'{a}}r, and
Girshick]{MaskRCNN}
He, K.; Gkioxari, G.; Doll{\'{a}}r, P.; Girshick, R.B.
\newblock Mask {R-CNN}.
\newblock {\em arXiv} {\bf 2017}, arXiv:1703.06870.

\end{thebibliography}
\end{document}